\documentclass{article} 
\usepackage{iclr2027_conference,times}
\iclrfinalcopy

\usepackage{amsmath,amsfonts,bm}

\def\eqref#1{equation~\ref{#1}}

\def\1{\bm{1}}

\DeclareMathAlphabet{\mathsfit}{\encodingdefault}{\sfdefault}{m}{sl}
\SetMathAlphabet{\mathsfit}{bold}{\encodingdefault}{\sfdefault}{bx}{n}

\usepackage{hyperref}
\usepackage[capitalize,noabbrev]{cleveref}
\usepackage{url}
\usepackage{algorithm}      
\usepackage{algpseudocode} 
\usepackage{amsmath,amssymb}
\usepackage{enumitem}
\usepackage{xspace}
\usepackage{booktabs}
\usepackage{tabularx}
\usepackage{array}
\usepackage{fvextra}
\usepackage{graphicx}
\usepackage{wrapfig}
\usepackage{adjustbox}
\usepackage[table]{xcolor}
\usepackage{subcaption}
\usepackage{fontawesome}
\usepackage{caption}
\usepackage{makecell}
\usepackage{needspace}
\usepackage[titles]{tocloft}
\usepackage[most]{tcolorbox}
\tcbuselibrary{skins}
\tcbset{
  aibox/.style={
    width=0.98\linewidth,
    top=8pt,
    bottom=4pt,
    colback=blue!6!white,
    colframe=blue!50!black,
    colbacktitle=blue!50!black,
    enhanced,
    center,
    attach boxed title to top left={yshift=-0.1in,xshift=0.15in},
    boxed title style={boxrule=0pt,colframe=white,},
  }
}
\newtcolorbox{AIbox}[2][]{aibox,title=#2,#1}

\newtcolorbox{promptbox}[1]{
  enhanced,
  breakable,
  boxrule = 1pt,
  fontupper = \scriptsize,
  fonttitle = \bf\color{white},
  arc = 5pt,
  rounded corners,
  colback=blue!4!white,
  colframe=blue!50!black,
  colbacktitle=blue!50!black,
  title = #1,
  left=2pt %
}

\title{\textsc{FrugalEvo}: Towards Cost-Aware LLM-Guided Program Evolution} 

\author{%
Hui Chen$^{1}$\thanks{Correspondence to: \texttt{\href{mailto:hui.chen@nus.edu.sg}{hui.chen@nus.edu.sg}}}
\quad Xuan Qi$^{2}$
\quad James Xu Zhao$^{1}$
\quad Zhaopeng Feng$^{1}$
\quad Shilong Liu$^{3}$\vspace{0.1em}\\
{\bfseries Kuang Xu$^{4}$
\quad Pang Wei Koh$^{2}$
\quad Bryan Hooi$^{1}$}\vspace{0.6em}\\
$^{1}$ National University of Singapore \quad $^{2}$ University of Washington \quad $^{3}$ Princeton University\vspace{0.1em}\\
$^{4}$ Stanford University\vspace{0.4em}\\
[0.25em]\mbox{\faGithub\ Code: \texttt{\href{https://github.com/chchenhui/frugalevo}{https://github.com/chchenhui/frugalevo}}}
}

\newcommand{\modelname}{\texttt{FrugalEvo}\xspace}
\newcommand{\metricname}{BA-AUC\xspace}

\begin{document}
\addtocontents{toc}{\protect\setcounter{tocdepth}{0}}

\maketitle

\begin{abstract}
LLM-guided evolutionary methods, such as AlphaEvolve, have emerged as powerful approaches for challenging computational optimization problems, such as circle packing. However, prior work typically optimizes performance gain over a fixed number of iterations. We argue that practical optimization should maximize \textbf{gain per unit cost}. To this end, we propose \texttt{FrugalEvo}, a cost-aware evolutionary framework where a stronger, higher-cost LLM explores solution strategies, and a cheaper LLM implements them and iteratively refines the resulting code. We also design a cache-efficient evolution process, where our harness and prompts maximize the sharing of prefixes across different evolution steps, to improve cache reuse. To measure solution quality throughout a fixed cost budget, we introduce Budget-Aware Area Under the Curve (BA-AUC), defined as the area under the best-so-far evaluation score curve over cumulative LLM cost, up to the budget. Across 10 mathematical and systems optimization tasks, \texttt{FrugalEvo} matches or surpasses state-of-the-art baselines, including OpenEvolve, ShinkaEvolve, AdaEvolve, and EvoX, in final solution quality and achieves higher BA-AUC on 9 tasks. It also achieves higher average performance than these baselines on 10 algorithmic optimization tasks from ALE-Bench-Lite. Notably, on circle packing, \texttt{FrugalEvo} achieves new state-of-the-art performance with GPT-5.6 Terra and Luna for only \$1.68 and with GLM-5.3 and its Flash variant for only \$0.55, matching or surpassing all baselines, including multi-agent methods such as CORAL and SwarmResearch, which cost approximately \$50 on average.
\end{abstract}

\begin{figure*}[ht]
  \centering
  \begin{subfigure}[b]{0.35\textwidth}
    \centering
    \includegraphics[width=\linewidth]{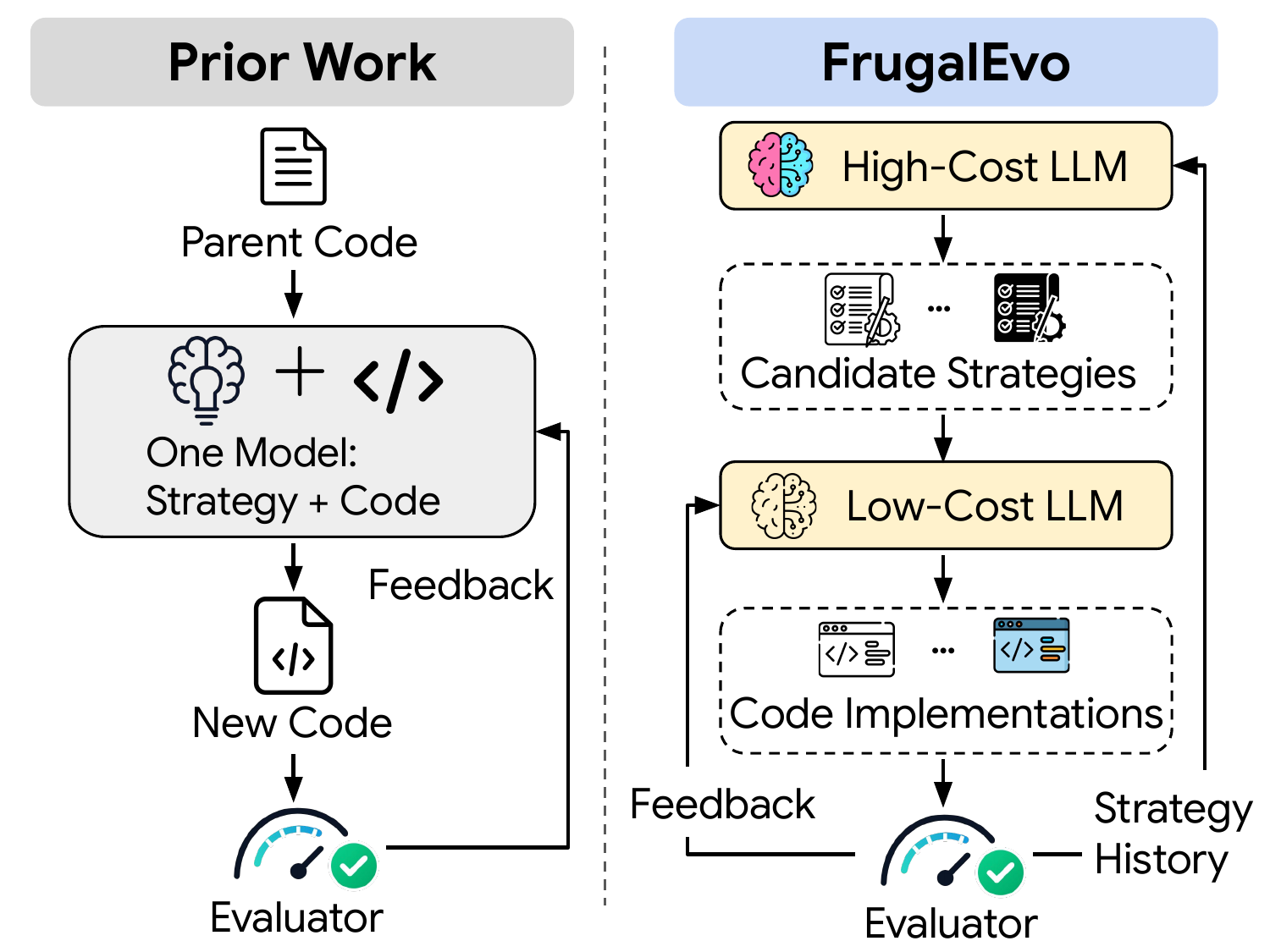}
    \caption{Comparison of frameworks.}
    \label{fig:framework_comparison}
  \end{subfigure}%
  \hspace{1mm}
  \begin{subfigure}[b]{0.32\textwidth}
    \centering
    \includegraphics[width=\linewidth]{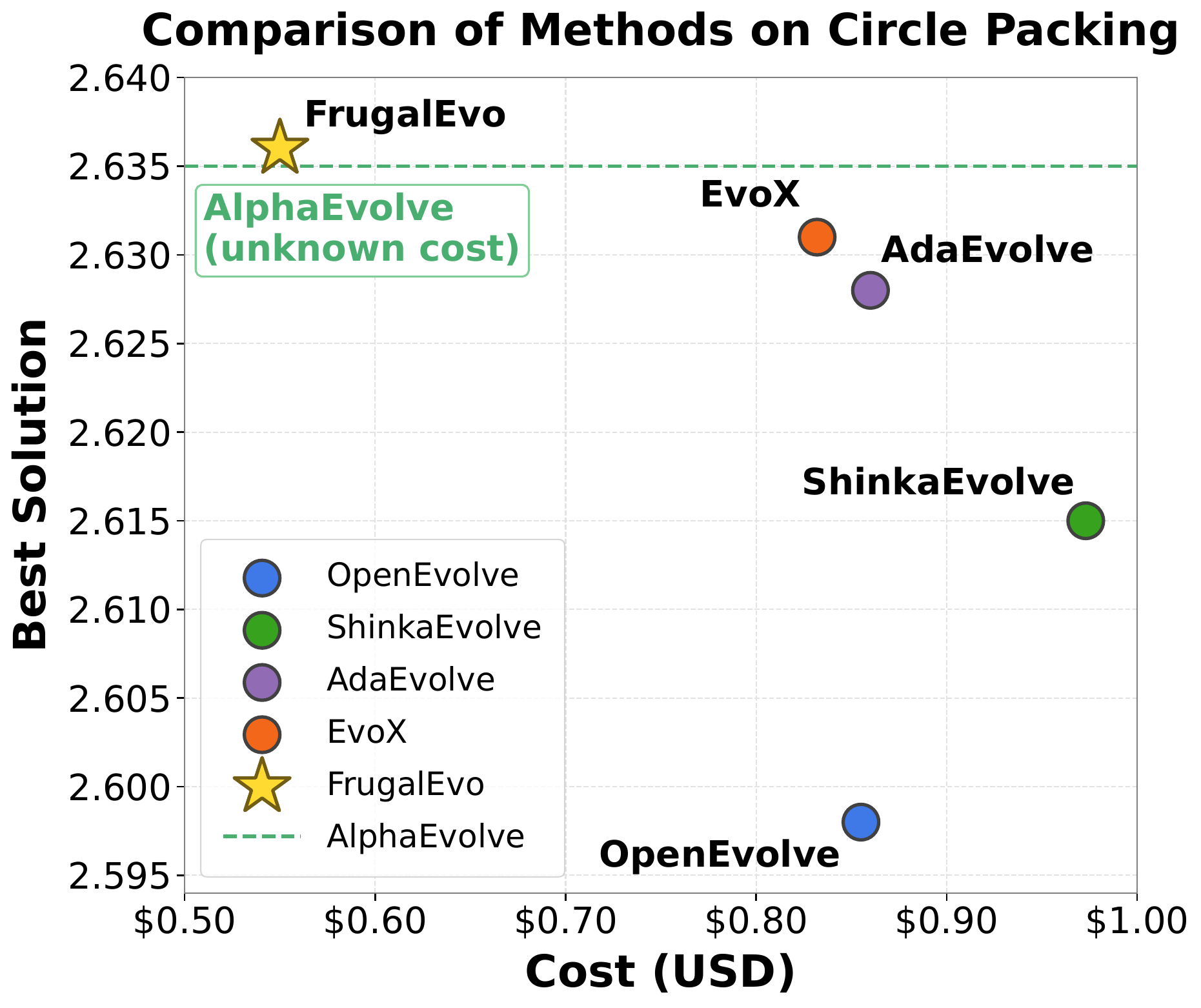}
    \caption{Solution quality vs. cost.}
    \label{fig:quality-cost}
  \end{subfigure}%
  \hspace{1mm}
  \begin{subfigure}[b]{0.30\textwidth}
    \centering
    \includegraphics[width=\linewidth]{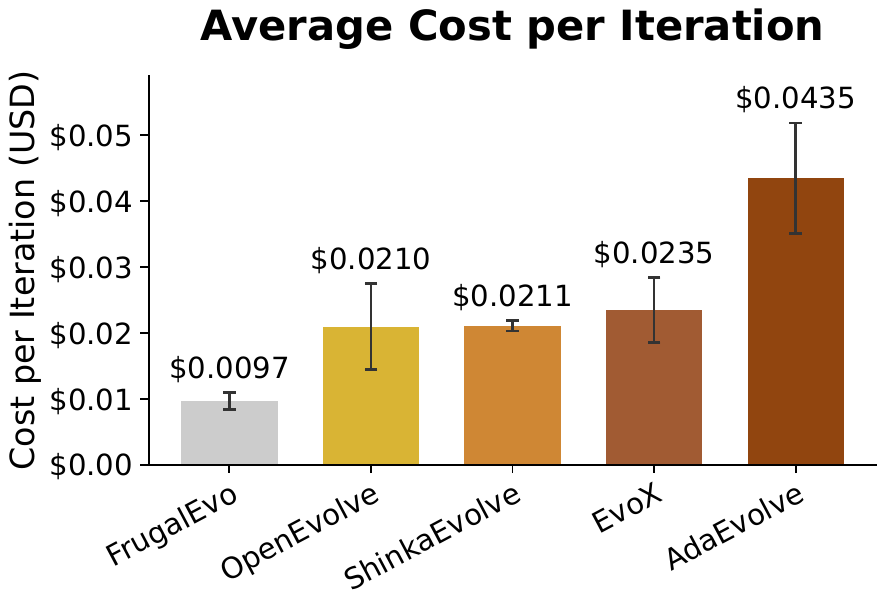}
    \caption{Average cost per iteration.}
    \label{fig:cost-iteration}
  \end{subfigure}
  \caption{High-level overview of \modelname{}. (a) compares the \modelname{}
framework with those of prior work. (b) and (c) evaluate methods on the circle
packing task: (b) plots solution quality against cumulative API cost, and (c)
reports the average cost per iteration. \modelname{} pairs GLM 5.3 with the
cheaper GLM 5.3 Flash, whereas all other methods use GLM 5.3 alone. Results are
averaged over three runs, each given a budget of \$1 (USD).}
  \label{fig:high_level_overview}
\end{figure*}

\section{Introduction}
Rapid advances in large language models (LLMs) have accelerated
scientific discovery across diverse domains~\citep{feng2026aletheia,liu2026lifescibench,wang2026frontierscience}. In particular, LLM-guided evolutionary methods, such as AlphaEvolve~\citep{novikov2025alphaevolve}, OpenEvolve~\citep{openevolve}, and ShinkaEvolve~\citep{lange2026shinkaevolve}, have shown promise in computational optimization problems, including mathematics~\citep{glazer2024frontiermath}, systems~\citep{cheng2025barbarians}, and algorithmic optimization~\citep{imajuku2026ale}. These methods typically maintain a pool of high-performing programs or solutions; they iteratively select a parent from this pool, prompt an LLM to propose a mutation, and then evaluate the resulting candidate. Progress is often measured as performance after a given number of evolutionary iterations~\citep{liu2026evox,cemri2026adaevolve}, with curves plotted against the number of LLM-generated candidates or evaluator calls.

However, focusing on performance after a fixed number of iterations
may overlook a key practical constraint in computational optimization:
the total cost. This cost may include API fees or GPU
compute costs for LLM inference, CPU compute costs for evaluator
execution, and other CPU/GPU resource costs associated with task
execution. On many challenging tasks, AI agents achieve better
performance with more inference-time computation, a phenomenon known
as \textit{test-time scaling}~\citep{snell2024scaling,brown2024large,muennighoff2025s1}. Consequently, higher performance alone
does not establish greater efficiency, as it may simply reflect
higher expenditure. We therefore argue that methods should be assessed
by their performance-cost trade-offs and their ability to maximize
solution quality under a fixed cost budget. This shifts the focus
from gain per iteration to \textbf{gain per unit cost}. To measure gain per unit cost, we introduce a new metric, Budget-Aware Area
Under the Curve (BA-AUC), based on the performance achieved at each cost budget.

Next, we propose \modelname, a cost-aware LLM-guided evolutionary framework for computational optimization. Since LLM inference often accounts for the largest proportion of search costs in this domain, compared with those of evaluator execution and other CPU/GPU operations, we focus on improving the cost efficiency of LLM inference. As shown in~\cref{fig:framework_comparison}, \modelname splits candidate program generation into two steps and uses two different models to carry them out: a strong, high-cost LLM explores diverse, promising design strategies, while a weaker, low-cost LLM implements them as executable programs and iteratively refines the resulting code.

Beyond model collaboration, we design a cache-efficient evolution
process for \modelname. Inference systems and APIs that support
prompt caching can reuse cached key-value (KV) states for identical
prompt prefixes, reducing latency and cost. To facilitate this reuse, we design our harness and prompts to maximize the sharing of prefixes across different evolution steps, improving cache reuse.

We evaluate \modelname on 20 tasks spanning mathematical, systems, and algorithmic optimization (\cref{sec:experiments}). With two model configurations---GPT-5.6 Terra and Luna~\citep{openai2026gpt56}, and GLM-5.3 and its Flash variant~\citep{zai2026glm53,zai2026glm53flash}---\modelname matches or outperforms OpenEvolve~\citep{openevolve}, ShinkaEvolve~\citep{lange2026shinkaevolve}, AdaEvolve~\citep{cemri2026adaevolve}, and EvoX~\citep{liu2026evox} in final solution quality and BA-AUC on 9 of 10 mathematical and systems optimization tasks. It also achieves the highest average performance across 10 algorithmic optimization tasks and matches or surpasses the reported AlphaEvolve~\citep{novikov2025alphaevolve} results on all four mathematical tasks with available comparisons. Notably, as shown in~\cref{fig:quality-cost}, \modelname achieves a sum of radii of 2.63599 on \textit{Circle Packing} using GLM-5.3 and its Flash variant for only \$0.55, surpassing all baselines, including multi-agent methods which cost approximately \$50 on average. Moreover, \cref{fig:cost-iteration} shows that \modelname substantially reduces the average cost per iteration.




In summary, our contributions include:
\begin{itemize}[leftmargin=16pt,nosep,itemsep=2pt]
   \item We argue that practical LLM-guided evolutionary search in computational optimization should focus on improving performance gain per unit cost. In this work, we introduce Budget-Aware Area Under the Curve (\metricname) to assess solution quality throughout a fixed LLM cost budget, complementing evaluation based on final scores.
    \item We propose \modelname, a cost-aware evolutionary framework
    that combines strategy exploration by a strong, high-cost model
    with code implementation and refinement by a low-cost model, and
    design a cache-efficient evolution process.
    \item We evaluate \modelname on 20 real-world computational optimization tasks. \modelname matches or surpasses all baselines, including OpenEvolve, ShinkaEvolve, AdaEvolve, and EvoX, in solution quality and achieves higher BA-AUC on 9 of 10 mathematical and systems optimization tasks. It also achieves the highest average performance on 10 algorithmic optimization tasks. 
\end{itemize}

\section{FrugalEvo: Cost-Aware Program Evolution}

\subsection{Problem Formulation}
Following prior work~\citep{novikov2025alphaevolve,romera2024mathematical},
we formulate LLM-guided evolutionary search as a program optimization process
under a fixed budget. 

\paragraph{Candidate Evaluation.} 
Let $\mathcal{X}$ denote the program space and let $x_0 \in \mathcal{X}$ be an initial program, which is an example given by the task. An evaluator evaluates each candidate $x \in \mathcal{X}$ and returns
a task-specific score and auxiliary feedback:
\begin{equation}
E(x) = (s(x), a(x)),
\qquad
E:\mathcal{X}\rightarrow\mathbb{R}\times\mathcal{A},
\end{equation}
where $s(x)\in\mathbb{R}$ is the task-specific score to be maximized and
$a(x)\in\mathcal{A}$ contains auxiliary feedback.

\paragraph{Program Database and Search Context.}
Evaluated programs and their results are stored in a program database.
Evaluating the initial program $x_0$ initializes the database as
$\mathcal{D}_0 = \{(x_0, s_0, a_0)\}$.
At iteration $t$, a search strategy $S$ constructs a search context
$C_S(\mathcal{D}_t)$ by combining the task description and analysis with a parent
program, sampled reference programs, their evaluation results, and
summaries of previous attempts retrieved from $\mathcal{D}_t$. Conditioned on this context, an LLM $G_{\mathrm{LLM}}$ generates a new candidate, which is then evaluated:
\begin{equation}
    x_{t+1}\sim
    G_{\mathrm{LLM}}\!\left(
        \cdot\mid C_S(\mathcal{D}_t)
    \right),
    \qquad
    E(x_{t+1})=(s_{t+1},a_{t+1}).
\end{equation}
The resulting record is added to the database as
$\mathcal{D}_{t+1}
=\mathcal{D}_t\cup\{(x_{t+1},s_{t+1},a_{t+1})\}$.

\paragraph{Optimization Goal.} As cost is a key factor in determining the practical utility of an approach, we focus on maximizing solution quality
\begin{equation}
    s^\star = \max_{(x,s,a)\in\mathcal{D}_B} s
\end{equation}
under a fixed cost budget $B$.

\paragraph{Budget-Aware AUC.}
\label{sec:ba_auc}
We introduce Budget-Aware Area Under the Curve (\metricname) to
measure how efficiently a method improves solution quality within
a fixed LLM cost budget. Let $q(c)$ denote the best-so-far evaluation score at cumulative LLM cost $c$. For a fixed cost budget
$B$, we define
\begin{equation}
    \operatorname{BA\text{-}AUC}(B)
    =
    \int_{0}^{B} q(c)\,\mathrm{d}c.
    \label{eq:ba_auc}
\end{equation}

To compare all methods over the same cost interval $[0,B]$, we hold the best-so-far evaluation score constant up to $B$ if a run ends before reaching the
budget, and truncate the curve at $B$ if the run exceeds it.

\begin{figure*}[!t]
    \centering
    \includegraphics[width=1.0\linewidth]{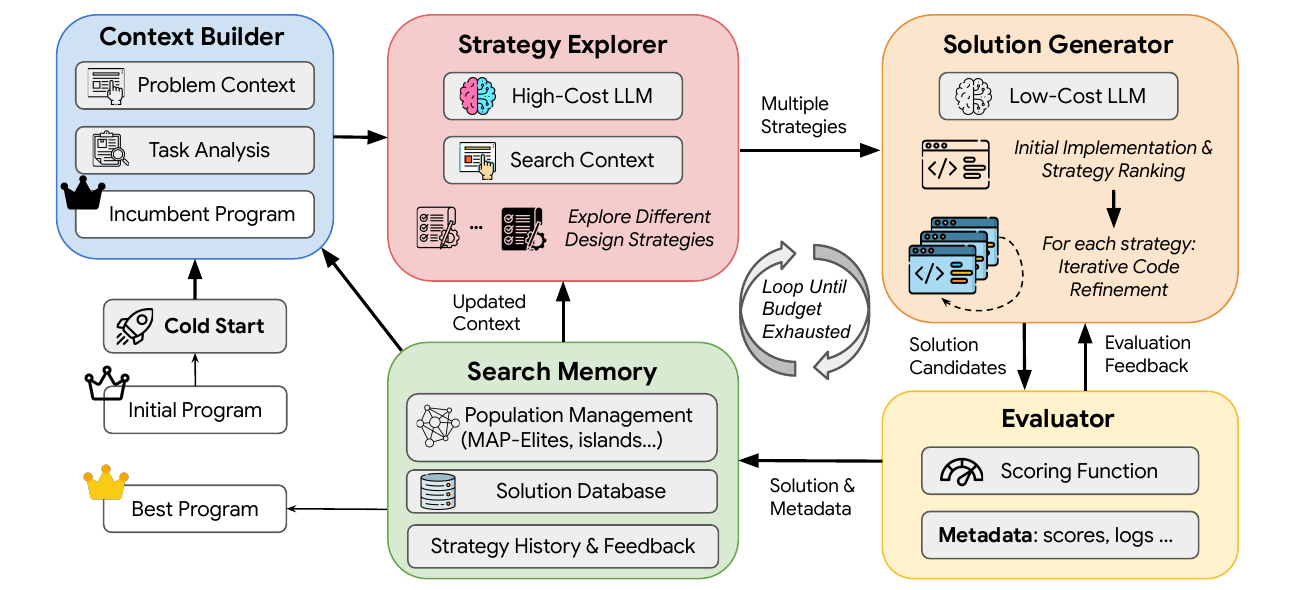}
    \caption{The \modelname framework. It comprises six main components: (1) a cold-start module, (2) a context builder, (3) a strategy explorer, (4) a solution generator, (5) an evaluator, and (6) a search memory. Our search memory uses island-based MAP-Elites~\citep{mouret2015illuminating,tanese1989distributed} algorithm.}
    \label{fig:framework_overview}
\end{figure*}

\subsection{Our Approach}

\paragraph{Overview.} \cref{fig:framework_overview} presents an overview of our approach, which proceeds as follows: (1) starting from an initial program, the cold start module establishes an ``incumbent''---the best-performing program found so far---as the parent code for strategy exploration and solution generation; (2) the context builder combines the problem context, task analysis, and incumbent program with information from search memory; (3) the strategy explorer uses a high-cost LLM to propose multiple design strategies; (4) the solution generator uses a low-cost LLM to implement and refine the proposed strategies; (5) the search memory updates the search context for the next round of strategy exploration; and (6) the process repeats until the budget is exhausted, returning the best program found. A detailed algorithm description is provided in Appendix~\ref{app:algorithm} and prompts of strategy explorer and solution generator can be found in Appendix~\ref{app:prompts}.

\paragraph{Cold Start.}
The strategy explorer generates design strategies based on a parent program. Using a weak initial program as the parent may lead to low-quality strategies and inefficient use of the cost budget. We therefore introduce a cold-start phase in which a low-cost LLM iteratively refines the initial program until the evaluation score no longer improves. The best resulting program becomes the initial incumbent and serves as the parent for the first round of strategy exploration.

\paragraph{Context Builder.}
The context builder organizes task information and search history to guide strategy exploration. At the start of evolution, the context builder uses a high-cost LLM to analyze the problem description and evaluator code to identify the optimization objective and evaluation constraints. It combines this information with the incumbent program established by cold start and its evaluation feedback to construct the initial search context. As the search progresses, it updates the context with the latest incumbent and attempted programs, strategies, and evaluation feedback retrieved from the search memory, guiding subsequent strategy exploration.

\paragraph{Cache-Efficient Prompt Construction.}
Specifically, to reduce input-token costs, we construct cache-efficient prompts by placing shared content at the beginning, enabling prefix reuse through API prompt caching.\footnote{A couple of LLM API providers support prompt caching, allowing repeated prompt prefixes to be reused at a lower input-token cost; e.g., OpenAI: \url{https://developers.openai.com/api/docs/guides/prompt-caching}.} To maximize prefix reuse between attempts, each prompt begins with fixed content (task instructions, output requirements, etc.). This is followed by information which changes less frequently, e.g., contextual examples, with island-best and elite programs appearing first before randomly sampled programs, lastly followed by frequently changing information, preserving the maximum amount of shared prefix across successive attempts. Additionally, feedback from each attempt is appended to the end of the prompt, allowing cache usage.

\paragraph{Strategy Explorer.}
Since algorithmic design can substantially affect solution quality, we use a strong, high-cost LLM to explore diverse design strategies. Given the search context, including the parent program, evaluation feedback, and previously explored strategies, it proposes multiple distinct strategies in a single API call. Each strategy provides a concrete plan for modifying the parent program, explaining what to change, why it may improve performance, and what computational limits to follow. The proposed strategies are then passed to the solution generator for implementation and refinement.

\paragraph{Solution Generator.}
The solution generator uses a low-cost LLM to implement and refine the proposed strategies. It first generates one trial implementation per strategy and ranks the strategies by their evaluation scores. It then follows this ranking to select strategies for further refinement. Each refinement round fixes one strategy and the current incumbent as the parent program, allowing up to \(M\) sequential implementation attempts. Each attempt is evaluated, and feedback from unsuccessful attempts informs subsequent prompts while the parent remains unchanged. As soon as an attempt improves the incumbent, the round ends, the improved program becomes the new incumbent, and another round begins with the same strategy and the updated parent. If a round ends without improvement, the generator moves to the next strategy.

\paragraph{Evaluator and Search Memory.}
The evaluator assesses each generated candidate program using task-specific scoring functions. Search memory uses island-based MAP-Elites~\citep{mouret2015illuminating,tanese1989distributed}
to maintain program diversity. It stores candidate programs and their evaluation results, along with proposed strategies, the results of strategy attempts, and failure feedback. The evolution process stops when the remaining budget cannot cover further LLM calls and returns the best program found.


\section{Experiments}
\label{sec:experiments}

\subsection{Experimental Setup}

\paragraph{Tasks.} We evaluate \modelname on 20 real-world optimization tasks, including 5 mathematics optimization problems~\citep{openevolve}, 5 systems optimization tasks from the ADRS benchmark~\citep{cheng2025barbarians}, and 10 algorithmic optimization problems from ALE-Bench-Lite~\citep{imajuku2026ale}. Benchmark details are provided in Appendix~\ref{app:task_description}.

\paragraph{Baselines.}
We compare \modelname against strong LLM-guided evolutionary methods,
including OpenEvolve~\citep{openevolve}, ShinkaEvolve~\citep{lange2026shinkaevolve}, AdaEvolve~\citep{cemri2026adaevolve}, and EvoX~\citep{liu2026evox}. For mathematical tasks, we additionally report state-of-the-art results from AlphaEvolve~\citep{novikov2025alphaevolve}.

\paragraph{Model Configurations and Budget.} We use two model configurations in our evaluation. In the first configuration, \modelname uses GPT-5.6 Terra~\citep{openai2026gpt56} for strategy generation and GPT-5.6 Luna~\citep{openai2026gpt56} for code implementation and refinement, while all four baselines use GPT-5.6 Terra. Each run receives an LLM cost budget of \$2 per run for mathematical optimization tasks and \$1 for systems and algorithmic optimization tasks. In the second configuration, \modelname uses GLM-5.3~\citep{zai2026glm53} for strategy generation and GLM-5.3 Flash~\citep{zai2026glm53flash} for implementation and refinement, while other baselines use GLM-5.3. The corresponding budgets are \$1 for mathematical tasks and \$0.5 for systems optimization tasks. Model Pricing details are provided in Appendix~\ref{app:model_pricing}.

\subsection{Main Results}
\paragraph{Results on Math Optimization Tasks.}
\label{sec:math_result}
\definecolor{bettergreen}{rgb}{0.9,1.0,0.9}

\begin{table*}[!t]
\centering
\setlength{\tabcolsep}{4pt}
\renewcommand{\arraystretch}{1.15}

\caption{\textbf{Results on mathematical optimization problems.} We report mean $\pm$ sample standard deviation and best scores over three independent runs. ``($\uparrow$)'' means the higher the better, while ``($\downarrow$)'' means the lower the better. \textbf{Bold} indicates the best result for each metric
within each task and model configuration. Light green highlights the best performance within each task across both model configurations.}
\label{tab:math_results}

\resizebox{0.9\linewidth}{!}{%
\begin{tabular}{lcccccc}
\toprule
& \multicolumn{3}{c}{\textbf{GPT-5.6 Terra/Luna (\$2)}}
& \multicolumn{3}{c}{\textbf{GLM-5.3/Flash (\$1)}} \\
\cmidrule(lr){2-4}\cmidrule(lr){5-7}
\textbf{Method}
& Avg. & Best & BA-AUC@2
& Avg. & Best & BA-AUC@1 \\
\midrule

\multicolumn{7}{l}{\textit{Circle Packing} ($\uparrow$)
\quad AlphaEvolve: $2.635000$} \\
OpenEvolve
& $2.634206_{\pm 0.0015}$
& $2.635983$
& $1.9709_{\pm 0.0007}$
& $2.555418_{\pm 0.0321}$
& $2.598002$
& $0.9438_{\pm 0.0157}$ \\
ShinkaEvolve
& $2.630981_{\pm 0.0003}$
& $2.631285$
& $1.9653_{\pm 0.0098}$
& $2.504844_{\pm 0.0837}$
& $2.615012$
& $0.9280_{\pm 0.0259}$ \\
AdaEvolve
& $2.632514_{\pm 0.0031}$
& $2.634305$
& $1.9653_{\pm 0.0039}$
& $2.618077_{\pm 0.0088}$
& $2.628041$
& $0.9375_{\pm 0.0162}$ \\
EvoX
& $2.635420_{\pm 0.0010}$
& $2.635983$
& $1.9549_{\pm 0.0228}$
& $2.599142_{\pm 0.0434}$
& $2.630972$
& $0.9497_{\pm 0.0337}$ \\
FrugalEvo
& \cellcolor{bettergreen}$\mathbf{2.635989}_{\pm \mathbf{0.0000}}$
& \cellcolor{bettergreen}$\mathbf{2.635996}$
& $\mathbf{1.9925}_{\pm \mathbf{0.0028}}$
& $\mathbf{2.635422}_{\pm \mathbf{0.0008}}$
& $\mathbf{2.635990}$
& $\mathbf{0.9939}_{\pm \mathbf{0.0039}}$ \\

\midrule
\multicolumn{7}{l}{\textit{Signal Processing} ($\uparrow$)} \\
OpenEvolve
& $0.55003_{\pm 0.0264}$
& $0.58045$
& $1.0802_{\pm 0.0308}$
& $0.66438_{\pm 0.0746}$
& $0.74245$
& $0.6297_{\pm 0.0662}$ \\
ShinkaEvolve
& $0.55604_{\pm 0.0301}$
& $0.59012$
& $1.0854_{\pm 0.0251}$
& $0.66246_{\pm 0.0475}$
& $0.71555$
& $0.6305_{\pm 0.0321}$ \\
AdaEvolve
& $0.72031_{\pm 0.0224}$
& $0.73677$
& $1.2520_{\pm 0.0225}$
& $0.67621_{\pm 0.0068}$
& $0.68281$
& $0.6315_{\pm 0.0086}$ \\
EvoX
& $0.64383_{\pm 0.0377}$
& $0.66574$
& $1.1706_{\pm 0.0428}$
& $0.67441_{\pm 0.0535}$
& $0.70925$
& $0.6401_{\pm 0.0609}$ \\
FrugalEvo
& $\mathbf{0.77791}_{\pm \mathbf{0.0145}}$
& \cellcolor{bettergreen}$\mathbf{0.78912}$
& $\mathbf{1.4819}_{\pm \mathbf{0.0308}}$
& \cellcolor{bettergreen}$\mathbf{0.78404}_{\pm \mathbf{0.0034}}$
& $\mathbf{0.78641}$
& $\mathbf{0.7539}_{\pm \mathbf{0.0068}}$ \\

\midrule
\multicolumn{7}{l}{\textit{Heilbronn Convex (13)} ($\uparrow$)
\quad AlphaEvolve: $0.03094$} \\
OpenEvolve
& $0.02314_{\pm 0.0024}$
& $0.02553$
& $1.3757_{\pm 0.1146}$
& $0.02854_{\pm 0.0021}$
& $0.03077$
& $0.8666_{\pm 0.0498}$ \\
ShinkaEvolve
& $0.02450_{\pm 0.0028}$
& $0.02769$
& $1.4499_{\pm 0.1325}$
& $0.02453_{\pm 0.0028}$
& $0.02758$
& $0.6531_{\pm 0.1748}$ \\
AdaEvolve
& $0.02349_{\pm 0.0011}$
& $0.02470$
& $1.4224_{\pm 0.0348}$
& $0.02797_{\pm 0.0021}$
& $0.03024$
& $0.8371_{\pm 0.0950}$ \\
EvoX
& $0.02248_{\pm 0.0002}$
& $0.02262$
& $1.3840_{\pm 0.0555}$
& $0.02896_{\pm 0.0008}$
& $0.02985$
& $0.8279_{\pm 0.0208}$ \\
FrugalEvo
& $\mathbf{0.02751}_{\pm \mathbf{0.0005}}$
& $\mathbf{0.02781}$
& $\mathbf{1.7405}_{\pm \mathbf{0.0393}}$
& \cellcolor{bettergreen}$\mathbf{0.03079}_{\pm \mathbf{0.0002}}$
& \cellcolor{bettergreen}$\mathbf{0.03094}$
& $\mathbf{0.9614}_{\pm \mathbf{0.0073}}$ \\

\midrule
\multicolumn{7}{l}{\textit{Heilbronn Triangle} ($\uparrow$)
\quad AlphaEvolve: $0.03653$} \\
OpenEvolve
& $0.03417_{\pm 0.0034}$
& \cellcolor{bettergreen}$\mathbf{0.03653}$
& $1.7517_{\pm 0.1833}$
& $0.03139_{\pm 0.0042}$
& $0.03598$
& $0.7256_{\pm 0.1359}$ \\
ShinkaEvolve
& $0.03030_{\pm 0.0004}$
& $0.03061$
& $1.4825_{\pm 0.0819}$
& $0.03051_{\pm 0.0010}$
& $0.03168$
& $0.7561_{\pm 0.0255}$ \\
AdaEvolve
& $0.03077_{\pm 0.0019}$
& $0.03273$
& $1.5951_{\pm 0.0485}$
& $0.03337_{\pm 0.0010}$
& $0.03456$
& $0.8192_{\pm 0.0505}$ \\
EvoX
& $0.03188_{\pm 0.0041}$
& \cellcolor{bettergreen}$\mathbf{0.03653}$
& $1.6444_{\pm 0.2031}$
& \cellcolor{bettergreen}$\mathbf{0.03486}_{\pm \mathbf{0.0016}}$
& $0.03617$
& $0.8316_{\pm 0.0260}$ \\
FrugalEvo
& $\mathbf{0.03480}_{\pm \mathbf{0.0008}}$
& $0.03568$
& $\mathbf{1.8733}_{\pm \mathbf{0.0132}}$
& $0.03463_{\pm 0.0017}$
& \cellcolor{bettergreen}$\mathbf{0.03653}$
& $\mathbf{0.8659}_{\pm \mathbf{0.0893}}$ \\

\midrule
\multicolumn{7}{l}{\textit{Min-Max-3} ($\downarrow$)
\quad AlphaEvolve: $4.165850$} \\
OpenEvolve
& \cellcolor{bettergreen}$\mathbf{4.165783}_{\pm \mathbf{0.0000}}$
& \cellcolor{bettergreen}$\mathbf{4.165783}$
& $1.9602_{\pm 0.0025}$
& $4.168021_{\pm 0.0039}$
& \cellcolor{bettergreen}$\mathbf{4.165783}$
& $0.9740_{\pm 0.0180}$ \\
ShinkaEvolve
& $4.166404_{\pm 0.0011}$
& \cellcolor{bettergreen}$\mathbf{4.165783}$
& $1.9356_{\pm 0.0096}$
& $4.167856_{\pm 0.0019}$
& \cellcolor{bettergreen}$\mathbf{4.165783}$
& $0.9737_{\pm 0.0043}$ \\
AdaEvolve
& \cellcolor{bettergreen}$\mathbf{4.165783}_{\pm \mathbf{0.0000}}$
& \cellcolor{bettergreen}$\mathbf{4.165783}$
& $1.9464_{\pm 0.0117}$
& $4.165784_{\pm 0.0000}$
& \cellcolor{bettergreen}$\mathbf{4.165783}$
& $0.9820_{\pm 0.0079}$ \\
EvoX
& \cellcolor{bettergreen}$\mathbf{4.165783}_{\pm \mathbf{0.0000}}$
& \cellcolor{bettergreen}$\mathbf{4.165783}$
& $1.9419_{\pm 0.0020}$
& $4.166583_{\pm 0.0014}$
& \cellcolor{bettergreen}$\mathbf{4.165783}$
& $0.9798_{\pm 0.0051}$ \\
FrugalEvo
& \cellcolor{bettergreen}$\mathbf{4.165783}_{\pm \mathbf{0.0000}}$
& \cellcolor{bettergreen}$\mathbf{4.165783}$
& $\mathbf{1.9769}_{\pm \mathbf{0.0159}}$
& \cellcolor{bettergreen}$\mathbf{4.165783}_{\pm \mathbf{0.0000}}$
& \cellcolor{bettergreen}$\mathbf{4.165783}$
& $\mathbf{0.9956}_{\pm \mathbf{0.0004}}$ \\

\bottomrule
\end{tabular}%
}
\end{table*}

We evaluate five tasks spanning geometric packing, extremal geometry,
distance-ratio minimization, and multi-objective signal filtering.
As shown in~\cref{tab:math_results}, \modelname achieves the highest
mean BA-AUC on all five tasks under both model configurations. It also
achieves the best or comparable mean performance on all five tasks
using GPT-5.6 Terra and Luna and four using GLM-5.3 and the Flash
variant. Compared to the strong reported results in the closed-source AlphaEvolve, \modelname
matches or exceeds AlphaEvolve on four tasks, including \textit{Circle Packing}, \textit{Heilbronn Convex (13)}, \textit{Heilbronn Triangle}, and \textit{Min-Max-3}.

\begin{figure}[!t]
    \centering
    \includegraphics[width=0.9\textwidth]{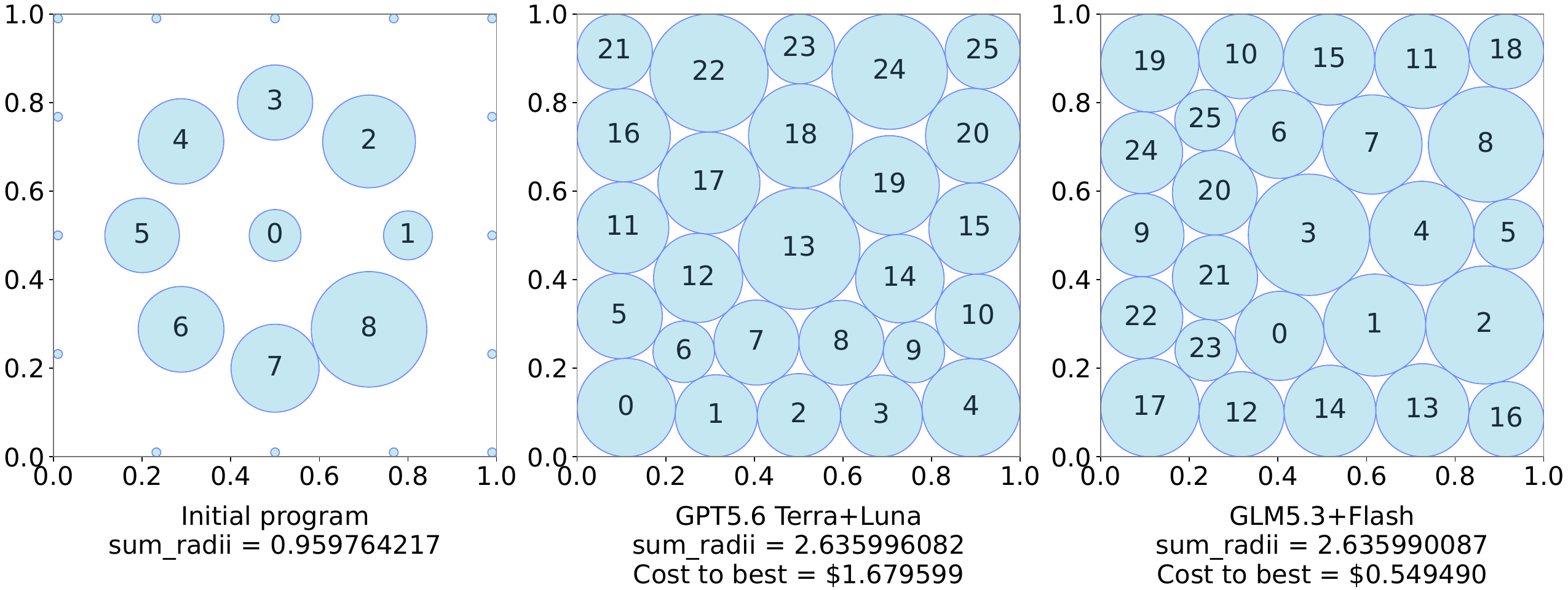}
    \caption{Visualization of \modelname solutions on the circle packing task. \textit{Left}: initial program; \textit{Middle}: \modelname solution generated by GPT-5.6 Terra and Luna; \textit{Right}: \modelname solution generated by GLM 5.3 and its Flash variant.}
    \label{fig:solution_visualization}
\end{figure}

\paragraph{Solution Analysis.} Beyond aggregate scores, \modelname discovers qualitatively distinct solutions across different tasks. On \textit{Circle Packing}, it finds a packing with a total radius of 2.635996 by evolving five-row layouts with asymmetric displacements, refining them through constrained SLSQP optimization and fixed-center radius linear programming, and applying small radius increases within the evaluator's tolerance. 

\cref{fig:solution_visualization} visualizes the solutions discovered by \modelname and compares them with the initial program on \textit{Circle Packing}. It shows that \modelname improves the
sum of radii on \textit{Circle Packing} from 0.95976 to 2.635996
using GPT-5.6 Terra and Luna for \$1.68, and to 2.635990 using
GLM-5.3 and its Flash variant for \$0.55. These results are better than or comparable
to those of CORAL~\citep{qu2026coral} and SwarmResearch~\citep{virk2026swarmresearch}, which cost approximately \$50 on average to achieve 2.635985 and 2.635996, respectively, demonstrating substantial cost savings.

Moreover, on \textit{Signal Processing}, \modelname reaches 0.78912 by adaptively blending Savitzky–Golay, Butterworth, and spectral filters while suppressing minor reversals. On \textit{Heilbronn Convex (13)}, \modelname matches AlphaEvolve through multistart hill climbing, linear programming refinement, and convex-hull area reduction. These results indicate the effectiveness of our approach: using a stronger model to explore strategies and a weaker model to iteratively refine solutions.

\paragraph{Results on Systems Optimization Tasks.}
\label{sec:sys_res}
\begin{table*}[!t]
\centering
\setlength{\tabcolsep}{4pt}
\renewcommand{\arraystretch}{1.15}
\caption{\textbf{Results on systems optimization problems.} We report mean $\pm$ sample standard deviation and best scores over three runs. ``($\uparrow$)'' means the higher the better, while ``($\downarrow$)'' means the lower the better. \textbf{Bold} indicates the best result for each metric within each task and model configuration. Light green highlights the best performance within each task across both model configurations.}
\label{tab:system_results}

\resizebox{0.9\linewidth}{!}{%
\begin{tabular}{lcccccc}
\toprule
& \multicolumn{3}{c}{\textbf{GPT-5.6 Terra/Luna (\$1)}}
& \multicolumn{3}{c}{\textbf{GLM-5.3/Flash (\$0.5)}} \\
\cmidrule(lr){2-4}\cmidrule(lr){5-7}
\textbf{Method}
& Avg. & Best & BA-AUC@1
& Avg. & Best & BA-AUC@0.5 \\
\midrule

\multicolumn{7}{l}{\textit{EPLB} ($\uparrow$)} \\
OpenEvolve
& $0.1346_{\pm 0.0097}$
& $0.1457$
& $0.1291_{\pm 0.0006}$
& $0.1352_{\pm 0.0101}$
& $0.1468$
& $0.0672_{\pm 0.0046}$ \\
ShinkaEvolve
& $0.1348_{\pm 0.0097}$
& $0.1460$
& $0.1314_{\pm 0.0044}$
& $0.1404_{\pm 0.0094}$
& $0.1469$
& $0.0678_{\pm 0.0027}$ \\
AdaEvolve
& $0.1295_{\pm 0.0004}$
& $0.1297$
& $0.1293_{\pm 0.0003}$
& $0.1291_{\pm 0.0012}$
& $0.1298$
& $0.0643_{\pm 0.0005}$ \\
EvoX
& $0.1355_{\pm 0.0099}$
& $0.1470$
& $0.1320_{\pm 0.0044}$
& $0.1294_{\pm 0.0006}$
& $0.1298$
& $0.0641_{\pm 0.0002}$ \\
FrugalEvo
& $\mathbf{0.1435}_{\pm \mathbf{0.0058}}$
& $\mathbf{0.1471}$
& $\mathbf{0.1408}_{\pm \mathbf{0.0047}}$
& \cellcolor{bettergreen}$\mathbf{0.1471}_{\pm \mathbf{0.0003}}$
& \cellcolor{bettergreen}$\mathbf{0.1473}$
& $\mathbf{0.0724}_{\pm \mathbf{0.0008}}$ \\

\midrule
\multicolumn{7}{l}{\textit{PRISM} ($\uparrow$)} \\
OpenEvolve
& $26.2307_{\pm 0.0400}$
& \cellcolor{bettergreen}$\mathbf{26.2560}$
& $25.9249_{\pm 0.0995}$
& \cellcolor{bettergreen}$\mathbf{26.2560}_{\pm \mathbf{0.0000}}$
& \cellcolor{bettergreen}$\mathbf{26.2560}$
& $12.9947_{\pm 0.0447}$ \\
ShinkaEvolve
& $26.2136_{\pm 0.0276}$
& $26.2332$
& $25.9399_{\pm 0.1104}$
& \cellcolor{bettergreen}$\mathbf{26.2560}_{\pm \mathbf{0.0000}}$
& \cellcolor{bettergreen}$\mathbf{26.2560}$
& $12.8018_{\pm 0.1138}$ \\
AdaEvolve
& $25.9493_{\pm 0.1707}$
& $26.0818$
& $25.1931_{\pm 0.4972}$
& $26.2539_{\pm 0.0035}$
& \cellcolor{bettergreen}$\mathbf{26.2560}$
& $12.8213_{\pm 0.1775}$ \\
EvoX
& $25.8464_{\pm 0.3368}$
& $26.2200$
& $24.9843_{\pm 0.5623}$
& $26.2217_{\pm 0.0594}$
& \cellcolor{bettergreen}$\mathbf{26.2560}$
& $12.7215_{\pm 0.2384}$ \\
FrugalEvo
& \cellcolor{bettergreen}$\mathbf{26.2560}_{\pm \mathbf{0.0000}}$
& \cellcolor{bettergreen}$\mathbf{26.2560}$
& $\mathbf{26.0695}_{\pm \mathbf{0.1844}}$
& \cellcolor{bettergreen}$\mathbf{26.2560}_{\pm \mathbf{0.0000}}$
& \cellcolor{bettergreen}$\mathbf{26.2560}$
& $\mathbf{13.1117}_{\pm \mathbf{0.0030}}$ \\

\midrule
\multicolumn{7}{l}{\textit{LLM-SQL} ($\uparrow$)} \\
OpenEvolve
& $0.7165_{\pm 0.0010}$
& $0.7175$
& $0.6263_{\pm 0.0258}$
& $0.7262_{\pm 0.0052}$
& $0.7296$
& $0.3349_{\pm 0.0044}$ \\
ShinkaEvolve
& $0.7204_{\pm 0.0058}$
& $0.7271$
& $0.6818_{\pm 0.0077}$
& $0.7236_{\pm 0.0066}$
& $0.7307$
& $0.3397_{\pm 0.0121}$ \\
AdaEvolve
& $0.7190_{\pm 0.0004}$
& $0.7193$
& $0.6502_{\pm 0.0047}$
& $0.7259_{\pm 0.0076}$
& $0.7321$
& $0.3374_{\pm 0.0064}$ \\
EvoX
& $0.7183_{\pm 0.0009}$
& $0.7192$
& $0.6426_{\pm 0.0319}$
& $0.7239_{\pm 0.0084}$
& $0.7302$
& $0.3370_{\pm 0.0080}$ \\
FrugalEvo
& $\mathbf{0.7248}_{\pm \mathbf{0.0085}}$
& $\mathbf{0.7313}$
& $\mathbf{0.7161}_{\pm \mathbf{0.0057}}$
& \cellcolor{bettergreen}$\mathbf{0.7336}_{\pm \mathbf{0.0009}}$
& \cellcolor{bettergreen}$\mathbf{0.7346}$
& $\mathbf{0.3566}_{\pm \mathbf{0.0121}}$ \\

\midrule
\multicolumn{7}{l}{\textit{Cloudcast} ($\downarrow$)} \\
OpenEvolve
& $618.103_{\pm 0.000}$
& $618.103$
& $0.1572_{\pm 0.0021}$
& $637.112_{\pm 17.314}$
& $627.100$
& $0.0702_{\pm 0.0063}$ \\
ShinkaEvolve
& $618.098_{\pm 0.004}$
& $618.093$
& $0.1568_{\pm 0.0027}$
& $725.286_{\pm 19.810}$
& $705.094$
& $0.0575_{\pm 0.0069}$ \\
AdaEvolve
& $618.098_{\pm 0.000}$
& $618.098$
& $\mathbf{0.1578}_{\pm \mathbf{0.0008}}$
& $785.177_{\pm 216.772}$
& $627.132$
& $0.0622_{\pm 0.0126}$ \\
EvoX
& $618.103_{\pm 0.000}$
& $618.103$
& $0.1572_{\pm 0.0015}$
& $630.026_{\pm 9.204}$
& $622.617$
& $0.0638_{\pm 0.0087}$ \\
FrugalEvo
& \cellcolor{bettergreen}$\mathbf{618.093}_{\pm \mathbf{0.016}}$
& \cellcolor{bettergreen}$\mathbf{618.075}$
& $0.1565_{\pm 0.0016}$
& $\mathbf{622.601}_{\pm \mathbf{4.506}}$
& $\mathbf{618.098}$
& $\mathbf{0.0776}_{\pm \mathbf{0.0013}}$ \\

\midrule
\multicolumn{7}{l}{\textit{Transaction Scheduling} ($\uparrow$)} \\
OpenEvolve
& $4091.92_{\pm 169.16}$
& $4237.29$
& $3893.72_{\pm 150.82}$
& $4052.41_{\pm 193.96}$
& $4291.85$
& $1971.43_{\pm 130.54}$ \\
ShinkaEvolve
& $4023.16_{\pm 203.94}$
& $4149.38$
& $3734.62_{\pm 111.63}$
& $3952.25_{\pm 240.14}$
& $4115.23$
& $1840.84_{\pm 71.76}$ \\
AdaEvolve
& $3858.51_{\pm 119.85}$
& $3984.06$
& $3720.02_{\pm 38.42}$
& $4047.19_{\pm 237.88}$
& $4255.32$
& $1955.27_{\pm 93.63}$ \\
EvoX
& $3913.30_{\pm 271.06}$
& $4219.41$
& $3674.87_{\pm 162.27}$
& $3975.20_{\pm 265.47}$
& $4184.10$
& $1800.25_{\pm 116.93}$ \\
FrugalEvo
& $\mathbf{4252.11}_{\pm \mathbf{110.39}}$
& \cellcolor{bettergreen}$\mathbf{4405.29}$
& $\mathbf{4133.73}_{\pm \mathbf{124.12}}$
& \cellcolor{bettergreen}$\mathbf{4288.01}_{\pm \mathbf{98.33}}$
& $\mathbf{4366.81}$
& $\mathbf{2107.07}_{\pm \mathbf{42.86}}$ \\

\bottomrule
\end{tabular}%
}
\end{table*}

\cref{tab:system_results} reports results on five real-world systems optimization tasks from the ADRS benchmark~\citep{cheng2025barbarians}. It shows that \modelname achieves the best
or comparable mean and best performance on all five tasks under both
model configurations, with the highest mean BA-AUC on four tasks.

Qualitatively, \modelname discovers effective approaches to systems optimization. On \textit{EPLB}, it achieves a score of 0.1473 with an expert load-balancing algorithm that combines greedy expert replication, load-aware block assignment, and pairwise swap refinement. On \textit{Transaction Scheduling}, it achieves a combined score of 4405.29 with a scheduler that combines multistart regret-guided insertion, exhaustive local window reordering, and transaction relocation to reduce makespan.

\paragraph{Results on Algorithmic Optimization Tasks.}

\begin{wrapfigure}{r}{0.37\linewidth}
\vspace{-1.1\baselineskip}
    \centering
    \includegraphics[width=\linewidth]{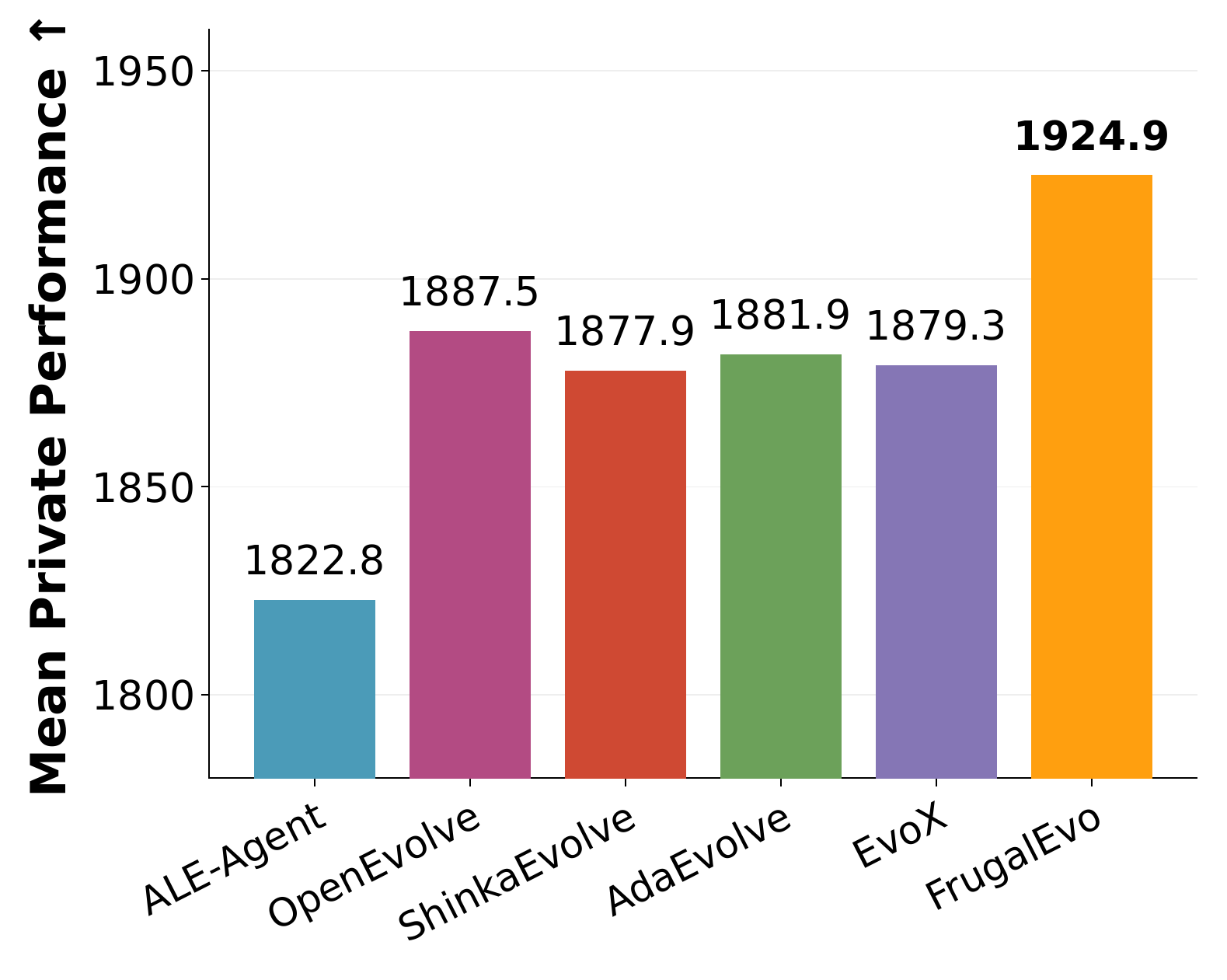}
    \caption{Average private performance across 10 ALE-Bench-Lite tasks. \modelname uses GPT-5.6 Terra and Luna and other baselines use GPT-5.6 Terra. The budget is \$1.}
    \label{fig:alebench_result}
    \vspace{-2.0\baselineskip}
\end{wrapfigure}

We evaluate \modelname on 10 algorithmic optimization tasks from ALE-Bench-Lite~\citep{imajuku2026ale}, derived from AtCoder
Heuristic Contests, and report the results in~\cref{fig:alebench_result}.
Following prior work~\citep{lange2026shinkaevolve}, all methods start
from solutions produced by ALE-Agent~\citep{imajuku2026ale} and are
evaluated using private scores.

As shown in~\cref{fig:alebench_result}, \modelname achieves the highest
score of 1924.9, outperforming strong baselines OpenEvolve
(1887.5) and AdaEvolve (1881.9). It discovers diverse algorithmic optimization solutions. On AHC016 (graph classification), \modelname constructs graphs from disjoint cliques and their complements, identifying noisy graphs by comparing the sorted numbers of neighbors per vertex with noise-adjusted expectations. On AHC027 (cleaning route), \modelname uses dirt accumulation rates, time since the last visit, and travel distances to plan routes that revisit areas requiring more frequent cleaning.

\paragraph{Performance-For-Cost Curves.}
\begin{figure*}[!t]
    \centering
    \includegraphics[width=1.0\textwidth]{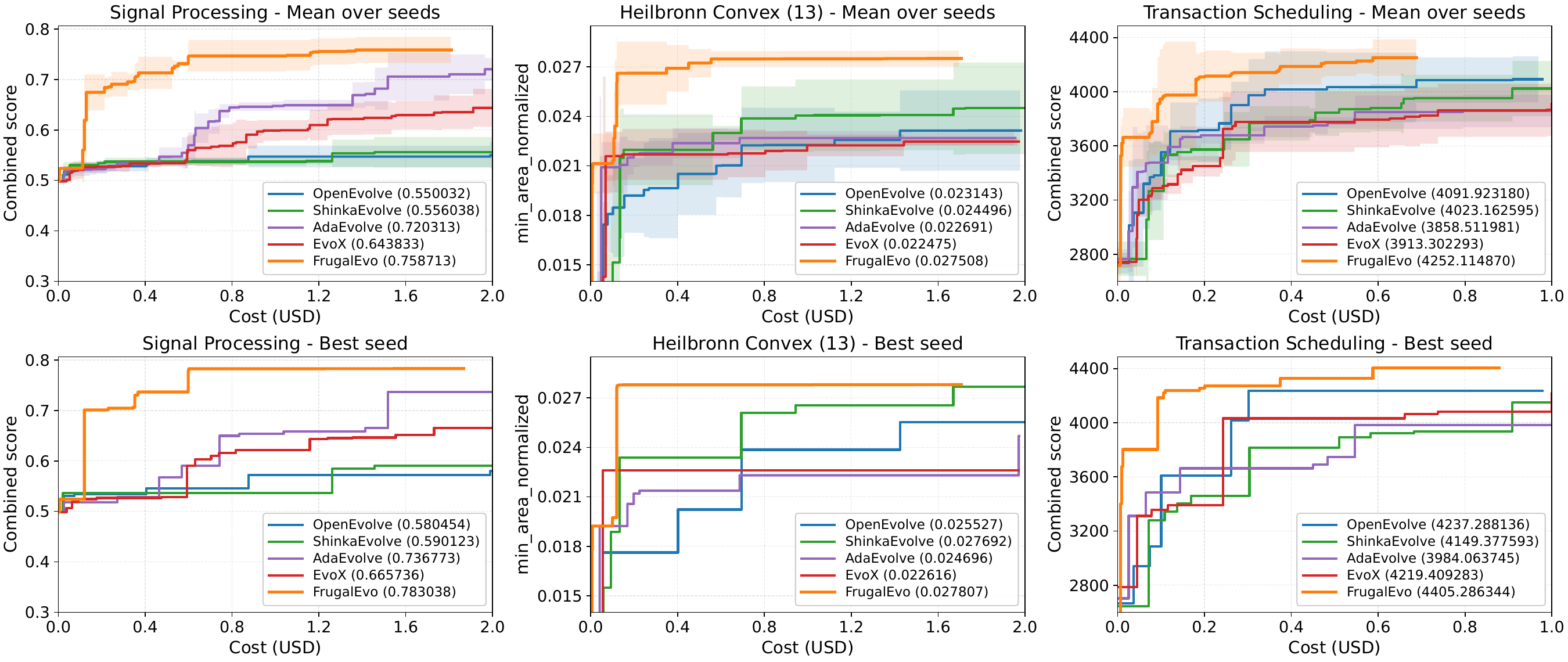}
    \caption{Best-so-far score curves on
signal processing, heilbronn convex, and transaction scheduling. \modelname uses GPT-5.6 Terra and
Luna, while the baselines use GPT-5.6 Terra. The top row shows mean trajectories
across three runs, and the bottom row shows the trajectory of the
best-performing run for each method. BA-AUC values are printed along with method names.}
    \label{fig:performance_cost}
\end{figure*}

To compare cost efficiency, we plot performance-for-cost curves of different methods on \textit{Signal Processing}, \textit{Heilbronn Convex (13)}, and \textit{Transaction Scheduling}. As shown in~\cref{fig:performance_cost}, \modelname establishes a clear performance advantage within the first \$0.2 on all three tasks and maintains it throughout the remaining search. Its higher mean BA-AUC values on all three tasks further quantify this advantage over the cost interval. These results indicate that \modelname achieves higher search efficiency within a given cost budget.

\subsection{Ablations}

In this section, we present ablation studies on key components of \modelname (model collaboration, cold start, and sequential feedback in code refinement) and the results are shown in~\cref{tab:ablation_results}.

\paragraph{Impact of Model Collaboration.}
Our framework assigns strategy exploration and solution generation
to different models to improve search efficiency.
As shown in~\cref{tab:ablation_results}, using either the high-cost
or low-cost model for both stages yields lower mean and best
performance than \modelname on both tasks under the same cost budget. In particular, using only the high-cost model leads to a larger drop in mean performance, possibly because implementation calls consume a substantial portion of the budget, leaving less for further exploration. Moreover, using only the low-cost model reduces the best final score, particularly on \textit{Signal Processing}. This may be because the low-cost model is less effective at identifying promising strategies. These results suggest that combining strategy exploration by a strong, expensive model with low-cost code implementation improves search efficiency under a fixed cost budget.

\begin{wraptable}{r}{0.39\textwidth}
\vspace{-1.1\baselineskip}
\centering
\caption{Ablation results with GPT-5.6 Terra and Luna.}
\label{tab:ablation_results}

\small

\begin{adjustbox}{max width=\linewidth}
\begin{tabular}{@{}lcc@{}}
\toprule
\textbf{Method} & Performance Avg. & Best \\
\midrule
\multicolumn{3}{@{}l}{\textit{Circle Packing} ($\uparrow$)} \\
FrugalEvo
& $\mathbf{2.635989}_{\pm \mathbf{0.0000}}$
& $\mathbf{2.635996}$ \\
- Terra + Terra
& $2.632278_{\pm 0.0063}$
& $2.635983$ \\
- Luna + Luna
& $2.634565_{\pm 0.0025}$
& $2.635983$ \\
w/o cold start
& $2.632417_{\pm 0.0031}$
& $2.635983$ \\
w/o seq feedback
& $2.633659_{\pm 0.0021}$
& $\mathbf{2.635996}$ \\
\midrule
\multicolumn{3}{@{}l}{\textit{Signal Processing} ($\uparrow$)} \\
FrugalEvo
& $\mathbf{0.77791}_{\pm \mathbf{0.0145}}$
& $\mathbf{0.78912}$ \\
- Terra + Terra
& $0.68534_{\pm 0.0352}$
& $0.70706$ \\
- Luna + Luna
& $0.74865_{\pm 0.0097}$
& $0.75973$ \\
w/o cold start
& $0.75576_{\pm 0.0240}$
& $0.77973$ \\
w/o seq feedback
& $0.76072_{\pm 0.0209}$
& $0.77438$ \\
\bottomrule
\end{tabular}
\end{adjustbox}
\vspace{-1.0\baselineskip}
\end{wraptable}



\paragraph{Impact of Cold Start and Sequential Feedback.}
We introduce a cold-start phase to provide a stronger initial
incumbent for strategy exploration. As shown in~\cref{tab:ablation_results},
removing this phase and using the initial program directly as the
parent reduces both mean and best performance on both tasks.
This suggests that refining the initial program with a low-cost model
is a cost-effective way to support high-quality strategy exploration.
Moreover, we also incorporate sequential feedback into code refinement in the solution generator.
Removing this feedback lowers mean performance and increases
the standard deviation across runs, suggesting that feedback from earlier
attempts helps the low-cost model refine programs more effectively
and improves search stability.
\section{Related Work}
\paragraph{LLM-Guided Evolutionary Methods.}
Recent work combining LLMs with evolutionary search has produced notable results~\citep{agrawal2026gepa,novikov2025alphaevolve} for mathematical~\citep{feng2026semi,feng2026towards,feng2026aletheia}, algorithmic~\citep{mang2025frontiercs}, and systems~\citep{cheng2025barbarians} optimization. For example, FunSearch~\citep{romera2024mathematical} combines LLM generation with automated evaluation to evolve mathematical solutions. AlphaEvolve~\citep{novikov2025alphaevolve} extends this approach from evolving individual functions to larger programs. Subsequently, OpenEvolve~\citep{openevolve} and CodeEvolve~\citep{assumpccao2025codeevolve} provide open-source frameworks, and ShinkaEvolve~\citep{lange2026shinkaevolve} introduces a sample-efficient evolution method. ThetaEvolve~\citep{wang2025thetaevolve} and TTT-Discover~\citep{yuksekgonul2026learning} apply Reinforcement Learning (RL) to improve the LLM generator at test time. More recently, AdaEvolve~\citep{cemri2026adaevolve} and EvoX~\citep{liu2026evox} adapt the search process, while SwarmResearch~\citep{virk2026swarmresearch} and CORAL~\citep{qu2026coral} coordinate multiple coding agents to improve search performance. These methods focus on performance gains over a fixed number of
iterations, while our work emphasizes gain per unit cost.

\paragraph{Cost-Efficient LLM Inference.}
Prior work explores LLM cascades~\citep{chen2023frugalgpt},
routing~\citep{ong2025routellm,feng2026llmrouter},
reflection~\citep{agrawal2026gepa,zhang2026agentic}, and
caching~\citep{zhang2026caching} to improve cost efficiency.
For instance, FrugalGPT~\citep{chen2023frugalgpt} learns model
cascades to balance cost and accuracy, while
RouteLLM~\citep{ong2025routellm} learns to route queries between
strong and weak models using preference data. For scientific discovery, SimpleTES~\citep{ye2026evaluation} post-trains an open-source model for long-horizon AI discovery by assigning each attempt the final outcome of the trajectory it helped produce. Meanwhile, Squeeze Evolve~\citep{maheswaran2026squeeze} coordinates multiple models for verifier-free evolution using models' own confidence signals, such as token log-probabilities. Our approach allocates model capability based on the characteristics of computational optimization tasks, without additional model training or access to token-level probabilities.

\section{Conclusion}
We introduce \modelname, a cost-aware LLM-guided evolutionary framework for computational optimization, that combines strategy exploration by a strong, high-cost model with code implementation and refinement by a low-cost model, and design a cache-efficient evolution process. To assess solution quality throughout a fixed cost budget, we propose Budget-Aware Area Under the Curve (BA-AUC). Across diverse real-world computational optimization tasks, \modelname matches or surpasses all baselines in solution quality and achieves higher BA-AUC on the majority of tasks.

\paragraph{Limitations and Future Work.} \modelname{} splits candidate program generation in computational optimization tasks into two steps and assigns them to two different models. This design is tailored to settings where both solution generation and evaluation take place entirely within a code environment, so that candidates can be executed and scored automatically. As a result, our method does not readily extend to scientific domains that require interaction with the physical world, such as biology, where evaluating a solution may involve wet-lab experiments that cannot be automated through code execution. Moreover, guiding the model to explore effective strategies relies on carefully crafted instructions. Our current instructions are manually designed for computational optimization tasks rather than automatically derived from each task. Finally, future work includes extending this approach to other domains that require interaction with the physical world, and develop methods that automatically generate strategy exploration instructions.

\section*{AI Use Statement}
In preparing this submission, we used LLMs as general-purpose
writing assistants and coding agents to implement the algorithm
and run experiments. The human authors were responsible for the
entire research process, including initial ideation, experimental
design, and data analysis. For the experiments, the authors first
developed an experimental plan and prepared the codebase and
datasets, then instructed coding agents to implement the proposed
method and execute the experiments. During writing, the authors
drafted each section, including all core ideas and technical
details, and used LLMs to improve clarity, coherence, and grammar.
All scientific claims and findings presented in this paper
originate from the authors, who take full responsibility for
the final content.

\section*{Ethics Statement}

Our work adheres to the ICLR Code of Ethics. The primary ethical considerations for this research involve the use of solutions generated by our algorithm across 20 optimization tasks. Although assessed by task evaluators, these solutions have not been validated by human experts and are intended for research reference only. 

\section*{Reproducibility Statement}

We are committed to ensuring the reproducibility of our research. To this end, we will publicly release all code, data, and supplementary materials upon publication.

\bibliography{iclr2027_conference}

@article{romera2024mathematical,
  title={Mathematical discoveries from program search with large language models},
  author={Romera-Paredes, Bernardino and Barekatain, Mohammadamin and Novikov, Alexander and Balog, Matej and Kumar, M Pawan and Dupont, Emilien and Ruiz, Francisco JR and Ellenberg, Jordan S and Wang, Pengming and Fawzi, Omar and others},
  journal={Nature},
  volume={625},
  number={7995},
  pages={468--475},
  year={2024},
  publisher={Nature Publishing Group UK London}
}

@article{novikov2025alphaevolve,
  title={Alphaevolve: A coding agent for scientific and algorithmic discovery},
  author={Novikov, Alexander and V{\~u}, Ng{\^a}n and Eisenberger, Marvin and Dupont, Emilien and Huang, Po-Sen and Wagner, Adam Zsolt and Shirobokov, Sergey and Kozlovskii, Borislav and Ruiz, Francisco JR and Mehrabian, Abbas and others},
  journal={arXiv preprint arXiv:2506.13131},
  year={2025}
}

@inproceedings{lange2026shinkaevolve,
  title={Shinkaevolve: Towards open-ended and sample-efficient program evolution},
  author={Lange, Robert and Imajuku, Yuki and Cetin, Edoardo},
  booktitle={International Conference on Learning Representations},
  volume={2026},
  pages={74026--74078},
  year={2026}
}

@article{wang2025thetaevolve,
  title={Thetaevolve: Test-time learning on open problems},
  author={Wang, Yiping and Su, Shao-Rong and Zeng, Zhiyuan and Xu, Eva and Ren, Liliang and Yang, Xinyu and Huang, Zeyi and He, Xuehai and Ma, Luyao and Peng, Baolin and others},
  journal={arXiv preprint arXiv:2511.23473},
  year={2025}
}

@misc{openevolve,
  title = {{OpenEvolve}: an open-source evolutionary coding agent},
  author = {Asankhaya Sharma},
  year = {2025},
  publisher = {GitHub},
  url = {https://github.com/algorithmicsuperintelligence/openevolve}
}

@article{assumpccao2025codeevolve,
  title={{CodeEvolve}: an open source evolutionary coding agent for algorithmic discovery and optimization},
  author={Assump{\c{c}}{\~a}o, Henrique and Ferreira, Diego and Campos, Leandro and Murai, Fabricio},
  journal={arXiv preprint arXiv:2510.14150},
  year={2025}
}

@article{liu2026evox,
  title={Evox: Meta-evolution for automated discovery},
  author={Liu, Shu and Agarwal, Shubham and Maheswaran, Monishwaran and Cemri, Mert and Li, Zhifei and Mang, Qiuyang and Naren, Ashwin and Boneh, Ethan and Cheng, Audrey and Pan, Melissa Z and others},
  journal={arXiv preprint arXiv:2602.23413},
  year={2026}
}

@inproceedings{agrawal2026gepa,
  title={Gepa: Reflective prompt evolution can outperform reinforcement learning},
  author={Agrawal, Lakshya A and Tan, Shangyin and Soylu, Dilara and Ziems, Noah and Khare, Rishi and Opsahl-Ong, Krista and Singhvi, Arnav and Shandilya, Herumb and Ryan, Michael J and Jiang, Meng and others},
  booktitle={International Conference on Learning Representations},
  volume={2026},
  pages={8479--8565},
  year={2026}
}

@article{cemri2026adaevolve,
  title={Adaevolve: Adaptive llm driven zeroth-order optimization},
  author={Cemri, Mert and Agrawal, Shubham and Gupta, Akshat and Liu, Shu and Cheng, Audrey and Mang, Qiuyang and Naren, Ashwin and Erdogan, Lutfi Eren and Sen, Koushik and Zaharia, Matei and others},
  journal={arXiv preprint arXiv:2602.20133},
  year={2026}
}

@article{virk2026swarmresearch,
  title={SwarmResearch: Orchestrating Coding Agents for Open-Ended Discovery},
  author={Virk, Yuvraj and Edds, Zack and Xia, Chunqiu Steven and Zhang, Lingming},
  journal={arXiv preprint arXiv:2607.02807},
  year={2026}
}

@article{qu2026coral,
  title={Coral: Towards autonomous multi-agent evolution for open-ended discovery},
  author={Qu, Ao and Zheng, Han and Zhou, Zijian and Yan, Yihao and Tang, Yihong and Ong, Shao Yong and Hong, Fenglu and Zhou, Kaichen and Jiang, Chonghe and Kong, Minwei and others},
  journal={arXiv preprint arXiv:2604.01658},
  year={2026}
}

@inproceedings{zhang2026agentic,
  title={Agentic context engineering: Evolving contexts for self-improving language models},
  author={Zhang, Qizheng and Hu, Changran and Upasani, Shubhangi and Ma, Boyuan and Hong, Fenglu and Kamanuru, Vamsidhar and Rainton, Jay and Wu, Chen and Ji, Mengmeng and Li, Hanchen and others},
  booktitle={International Conference on Learning Representations},
  volume={2026},
  pages={86069--86100},
  year={2026}
}

@article{feng2026llmrouter,
  title={{LLMRouter}: Unified Infrastructure for Developing, Evaluating, and Deploying LLM Routers},
  author={Feng, Tao and Yu, Fangxu and Zhang, Haozhen and Dai, Zhongjie and Yuan, Liangqi and Lei, Zijie and Zhang, Weizhi and Zhu, Kunlun and Yue, Haodong and Xuan, Keyang and others},
  journal={arXiv preprint arXiv:2608.06867},
  year={2026}
}

@inproceedings{ong2025routellm,
  title={Routellm: Learning to route llms from preference data},
  author={Ong, Isaac and Almahairi, Amjad and Wu, Vincent and Chiang, Wei-Lin and Wu, Tianhao and Gonzalez, Joseph E and Kadous, Mohammed and Stoica, Ion},
  booktitle={International Conference on Learning Representations},
  volume={2025},
  pages={34433--34448},
  year={2025}
}

@article{chen2023frugalgpt,
  title={Frugalgpt: How to use large language models while reducing cost and improving performance},
  author={Chen, Lingjiao and Zaharia, Matei and Zou, James},
  journal={arXiv preprint arXiv:2305.05176},
  year={2023}
}

@article{maheswaran2026squeeze,
  title={Squeeze evolve: Unified multi-model orchestration for verifier-free evolution},
  author={Maheswaran, Monishwaran and Lakhani, Leon and Zhou, Zhongzhu and Yang, Shijia and Wang, Junxiong and Hooper, Coleman and Hu, Yuezhou and Tiwari, Rishabh and Wang, Jue and Singh, Harman and others},
  journal={arXiv preprint arXiv:2604.07725},
  year={2026}
}

@article{snell2024scaling,
  title={Scaling llm test-time compute optimally can be more effective than scaling model parameters},
  author={Snell, Charlie and Lee, Jaehoon and Xu, Kelvin and Kumar, Aviral},
  journal={arXiv preprint arXiv:2408.03314},
  year={2024}
}

@article{brown2024large,
  title={Large language monkeys: Scaling inference compute with repeated sampling},
  author={Brown, Bradley and Juravsky, Jordan and Ehrlich, Ryan and Clark, Ronald and Le, Quoc V and R{\'e}, Christopher and Mirhoseini, Azalia},
  journal={arXiv preprint arXiv:2407.21787},
  year={2024}
}

@inproceedings{muennighoff2025s1,
  title={s1: Simple test-time scaling},
  author={Muennighoff, Niklas and Yang, Zitong and Shi, Weijia and Li, Xiang Lisa and Fei-Fei, Li and Hajishirzi, Hannaneh and Zettlemoyer, Luke and Liang, Percy and Cand{\`e}s, Emmanuel and Hashimoto, Tatsunori B},
  booktitle={Proceedings of the 2025 Conference on Empirical Methods in Natural Language Processing},
  pages={20286--20332},
  year={2025}
}

@article{ye2026evaluation,
  title={Evaluation-driven scaling for scientific discovery},
  author={Ye, Haotian and Lin, Haowei and Tang, Jingyi and Luo, Yizhen and Yang, Caiyin and Su, Chang and Thapa, Rahul and Yang, Rui and Liu, Ruihua and Li, Zeyu and others},
  journal={arXiv preprint arXiv:2604.19341},
  year={2026}
}

@article{yuksekgonul2026learning,
  title={Learning to discover at test time},
  author={Yuksekgonul, Mert and Koceja, Daniel and Li, Xinhao and Bianchi, Federico and McCaleb, Jed and Wang, Xiaolong and Kautz, Jan and Choi, Yejin and Zou, James and Guestrin, Carlos and others},
  journal={arXiv preprint arXiv:2601.16175},
  year={2026}
}

@article{mang2025frontiercs,
  title={Frontiercs: Evolving challenges for evolving intelligence},
  author={Mang, Qiuyang and Chai, Wenhao and Li, Zhifei and Mao, Huanzhi and Zhou, Shang and Du, Alexander and Li, Hanchen and Liu, Shu and Chen, Edwin and Wang, Yichuan and others},
  journal={arXiv preprint arXiv:2512.15699},
  year={2025}
}

@article{glazer2024frontiermath,
  title={Frontiermath: A benchmark for evaluating advanced mathematical reasoning in ai},
  author={Glazer, Elliot and Erdil, Ege and Besiroglu, Tamay and Chicharro, Diego and Chen, Evan and Gunning, Alex and Olsson, Caroline Falkman and Denain, Jean-Stanislas and Ho, Anson and Santos, Emily de Oliveira and others},
  journal={arXiv preprint arXiv:2411.04872},
  year={2024}
}

@article{imajuku2026ale,
  title={Ale-bench: A benchmark for long-horizon objective-driven algorithm engineering},
  author={Imajuku, Yuki and Horie, Kohki and Iwata, Yoichi and Aoki, Kensho and Takahashi, Naohiro and Akiba, Takuya},
  journal={Advances in Neural Information Processing Systems},
  volume={38},
  year={2026}
}

@article{cheng2025barbarians,
  title={Barbarians at the gate: How ai is upending systems research},
  author={Cheng, Audrey and Liu, Shu and Pan, Melissa and Li, Zhifei and Wang, Bowen and Krentsel, Alex and Xia, Tian and Cemri, Mert and Park, Jongseok and Yang, Shuo and others},
  journal={arXiv preprint arXiv:2510.06189},
  year={2025}
}

@techreport{openai2026gpt56,
  title={{GPT-5.6} System Card},
  author={{OpenAI}},
  year={2026},
  month={July},
  institution={OpenAI},
  url={https://deploymentsafety.openai.com/gpt-5-6/gpt-5-6.pdf}
}

@techreport{zai2026glm53,
  title={{GLM-5.3}: Frontier Coding with Emergent Cyber Capabilities},
  author={{Z.ai}},
  year={2026},
  month={August},
  institution={Z.ai},
  url={https://z.ai/blog/glm-5.3}
}

@techreport{zai2026glm53flash,
  title={{GLM-5.3-Flash}: Frontier Intelligence, Flash Cost},
  author={{Z.ai}},
  year={2026},
  month={August},
  institution={Z.ai},
  url={https://z.ai/blog/glm-5.3-flash}
}

@article{feng2026semi,
  title={Semi-Autonomous Mathematics Discovery with Gemini: A Case Study on the Erd{\H{o}}s Problems},
  author={Feng, Tony and Trinh, Trieu and Bingham, Garrett and Kang, Jiwon and Zhang, Shengtong and Kim, Sang-hyun and Barreto, Kevin and Schildkraut, Carl and Jung, Junehyuk and Seo, Jaehyeon and others},
  journal={arXiv preprint arXiv:2601.22401},
  year={2026}
}

@article{feng2026towards,
  title={Towards autonomous mathematics research},
  author={Feng, Tony and Trinh, Trieu H and Bingham, Garrett and Hwang, Dawsen and Chervonyi, Yuri and Jung, Junehyuk and Lee, Joonkyung and Pagano, Carlo and Kim, Sang-hyun and Pasqualotto, Federico and others},
  journal={arXiv preprint arXiv:2602.10177},
  year={2026}
}

@article{feng2026aletheia,
  title={Aletheia tackles FirstProof autonomously},
  author={Feng, Tony and Jung, Junehyuk and Kim, Sang-hyun and Pagano, Carlo and Gukov, Sergei and Tsai, Chiang-Chiang and Woodruff, David and Javanmard, Adel and Mokhtari, Aryan and Hwang, Dawsen and others},
  journal={arXiv preprint arXiv:2602.21201},
  year={2026}
}

@article{liu2026lifescibench,
  title={{LifeSciBench}: Evaluating Language Models on Realistic, Expert-Level Tasks in the Life Sciences},
  author={Liu, Amelia and Ho, Andrew and Droste, Anne Marie and Martin, David and Wong, Edmund and Zhou, Edward and Zhou, Isabelle and Park, Joshua and Jiao, Joy and Skelly, Katie-Rose and others},
  journal={bioRxiv},
  pages={2026--08},
  year={2026},
  publisher={Cold Spring Harbor Laboratory}
}

@article{wang2026frontierscience,
  title={{FrontierScience}: Evaluating AI's Ability to Perform Expert-Level Scientific Tasks},
  author={Wang, Miles and Lin, Robi and Hu, Kat and Jiao, Joy and Chowdhury, Neil and Chang, Ethan and Patwardhan, Tejal},
  journal={arXiv preprint arXiv:2601.21165},
  year={2026}
}

@article{mouret2015illuminating,
  title={Illuminating search spaces by mapping elites},
  author={Mouret, Jean-Baptiste and Clune, Jeff},
  journal={arXiv preprint arXiv:1504.04909},
  year={2015}
}

@book{tanese1989distributed,
  title={Distributed genetic algorithms for function optimization},
  author={Tanese, Reiko},
  year={1989},
  publisher={University of Michigan}
}

@article{zhang2026caching,
  title={Agentic plan caching: Test-time memory for fast and cost-efficient LLM agents},
  author={Zhang, Qizheng and Wornow, Michael and Olukotun, Kunle},
  journal={Advances in Neural Information Processing Systems},
  volume={38},
  pages={103270--103296},
  year={2026}
}
\bibliographystyle{iclr2027_conference}

\newpage
\appendix
\addtocontents{toc}{\protect\setcounter{tocdepth}{2}}
\noindent{\Large \textbf{Appendix}}

\makeatletter
\@starttoc{toc}
\makeatother

\vspace{10pt}

\section{Algorithm Description}
\label{app:algorithm}
\cref{alg:main_loop} describes our algorithm, which mainly contains four phases: (1) cold start, (2) strategy exploration, (3) initial code implementation and strategy ranking, and (4) iterative code refinement.
\begin{algorithm}[ht]
\caption{\modelname: Cost-aware LLM-guided program evolution}
\label{alg:main_loop}
\begin{algorithmic}[1]
\Require Budget $B$; task $d$; evaluator $E$; packages $\mathcal{P}$;
evaluated database $\mathcal{D}$
\Require High-cost LLM $\mathcal{G}_h$; low-cost LLM $\mathcal{G}_l$;
opening cap $N_0$; strategy count $K$; round cap $M$
\Require Consecutive-invalid limit $F$; tolerance $\epsilon$;
flags $I_0,I_{\mathrm{ref}}$
\State $\mathcal{H}\gets\varnothing$;
$x^\star\gets\arg\max_{x\in\mathcal{D}_{\mathrm{valid}}}s(x)$

\Statex \textbf{Phase I: Cold start}
\State $(\mathcal{D},x^\star)\gets
\Call{ColdStartAndCheapClimb}
{\mathcal{G}_l,\mathcal{D},x^\star,N_0,M,F,\epsilon,I_0,B}$
\Comment{Cheap search until stalled}
\State $A\gets\Call{AnalyzeOnce}{\mathcal{G}_h,d,E,\mathcal{P},B}$
\Comment{Cache evaluator and task analysis}
\While{\Call{BudgetAvailable}{B}}
    \Comment{New strategy portfolio}

    \Statex \textbf{Phase II: Strategy exploration}
    \State $\bar{x}\gets x^\star$;
    $\mathcal{C}\gets\Call{Context}{\mathcal{D},\bar{x}}$
    \Comment{Build search context}
    \State $(\{S_i\}_{i=1}^{k},x_{\mathrm{ref}})
    \gets\Call{Explore}
    {\mathcal{G}_h,\bar{x},\mathcal{C},\mathcal{H},
     A,K,I_{\mathrm{ref}},B}$
    \Comment{Propose $K$ strategies}
    \State $(\mathcal{D},x^\star,\bar{x},\mathcal{C},\mathcal{H})
    \gets\Call{ProcessReference}
    {x_{\mathrm{ref}},\mathcal{D},x^\star,\bar{x},
     \mathcal{C},\mathcal{H},\epsilon}$
    \Comment{Evaluate reference; use as shared parent if improved}

    \Statex \textbf{Phase III: Initial code implementation and strategy ranking}
    \State $(\mathcal{T},\mathbf{q},\mathcal{D},x^\star)
    \gets\Call{ParallelTrials}
    {\mathcal{G}_l,\{S_i\}_{i=1}^{k},\bar{x},
     \mathcal{C},\mathcal{D},x^\star,B}$
    \Comment{One trial per strategy; shared parent}
    \State $\mathcal{H}\gets\Call{RecordTrials}{\mathcal{H},\mathcal{T}}$;
    $\pi\gets\operatorname{StableArgsort}_{i}(-q_i)$
    \Comment{Rank strategies by trial scores}

    \Statex \textbf{Phase IV: Iterative code refinement}
    \For{$j=1,\ldots,k$}
        \While{\Call{BudgetAvailable}{B}}
            \State $b\gets s(x^\star)$
            \Comment{Cache the incumbent score}
            \State $(\mathcal{R},\mathcal{D},x^\star)
            \gets\Call{SequentialImprove}
            {\mathcal{G}_l,S_{\pi(j)},\mathcal{D},
             x^\star,M,F,\epsilon,B}$
            \Comment{Up to $M$ attempts; end round on improvement}
            \State $\Delta\gets s(x^\star)-b$;
            $\mathcal{H}\gets\Call{Record}
            {\mathcal{H},S_{\pi(j)},\mathcal{R},\Delta}$
            \Comment{Record global improvement}
            \If{$\Delta\le\epsilon$}
                \State \textbf{break}
                \Comment{Move to the next strategy}
            \EndIf
        \EndWhile
    \EndFor
\EndWhile
\State \Return $x^\star$
\end{algorithmic}
\end{algorithm}

\section{Benchmark Details}
\label{app:task_description}
We evaluate \modelname on 20 optimization problems spanning mathematical optimization (5 problems), systems optimization (5 problems), and algorithmic challenges (10 problems). 

\subsection{Mathematical Optimization}

\begin{table}[hb]
\centering
\caption{\textbf{Mathematical optimization tasks.} Summary of optimization tasks used to evaluate \modelname, including geometric optimization and signal processing problems.}
\label{tab:math_tasks}
\small
\renewcommand{\arraystretch}{1.15}
\begin{tabularx}{\linewidth}{@{}
  >{\raggedright\arraybackslash}p{0.27\linewidth}
  >{\raggedright\arraybackslash}p{0.17\linewidth}
  >{\raggedright\arraybackslash}X
@{}}
\toprule
\textbf{Problem} & \textbf{Objective} & \textbf{Description} \\
\midrule
Circle Packing (Square) & $\max \sum_{i=1}^{26} r_i$ &
Pack 26 circles with variable radii into the unit square, maximizing
the sum of their radii subject to non-overlap and containment constraints. \\

Signal Processing & Maximize composite score &
Filter noisy, non-stationary time series, balancing directional
reversals, recent and overall tracking errors, false reversals,
correlation with the clean signal, noise reduction, and run success
across five test signals. \\

Heilbronn Convex ($n=13$) & $\max A_{\min}/A_{\mathrm{hull}}$ &
Arrange 13 points in the plane to maximize the smallest triangle area
over all point triples, normalized by the area of the convex hull
of all 13 points. \\

Heilbronn Triangle ($n=11$) & $\max A_{\min}/A_T$ &
Place 11 points inside or on a fixed equilateral triangle of side
length one, maximizing the smallest triangle area over all point
triples, normalized by the container area $A_T=\sqrt{3}/4$. \\

Min-Max Distance Ratio (3D) &
$\max (d_{\min}/d_{\max})^2$ &
Arrange 14 points in three-dimensional Euclidean space to maximize
the squared ratio of the minimum to maximum pairwise Euclidean
distance, without a prescribed bounding region. \\
\bottomrule
\end{tabularx}
\end{table}

\subsection{Systems Optimization}

\begin{table}[hb]
\centering
\caption{\textbf{Systems optimization tasks.} Benchmarks drawn from ADRS-Bench~\citep{cheng2025barbarians}, capturing realistic optimization problems from production systems.}
\label{tab:sys_tasks}
\small
\renewcommand{\arraystretch}{1.15}
\begin{tabularx}{\linewidth}{@{}
  >{\raggedright\arraybackslash}p{0.27\linewidth}
  >{\raggedright\arraybackslash}p{0.17\linewidth}
  >{\raggedright\arraybackslash}X
@{}}
\toprule
\textbf{Problem} & \textbf{Objective} & \textbf{Description} \\
\midrule
EPLB (Expert Parallelism Load Balancer) & Maximize composite score &
Replicate and place MoE experts across GPUs using recorded expert
loads, balancing load across expert replicas while reducing
rebalancing and simulation runtime. \\

PRISM (GPU Model Placement) & Minimize max KV-cache pressure &
Assign models to GPUs under memory constraints, minimizing the
maximum ratio of SLO-weighted request load to memory remaining
for the KV cache. \\

LLM-SQL (Prefix Caching) & Maximize cache reuse and speed &
Optimize column orderings when serializing table rows into LLM
prompts, maximizing shared-prefix cache hits while minimizing
reordering runtime across five datasets. \\

Cloudcast (Multi-Cloud Broadcast) & Minimize total cost &
Construct partitioned broadcast routes from one source region to
multiple destinations, exploiting shared relay paths to reduce
egress and instance costs under heterogeneous bandwidths and
pricing. \\

Transaction Scheduling & Minimize makespan &
Order transactions with read--write and write--write conflicts
on shared data items to minimize total simulated makespan
across three workloads. \\
\bottomrule
\end{tabularx}
\end{table}

\subsection{Algorithmic Challenges}


\begin{table}[H]
\centering
\caption{\textbf{Algorithmic optimization tasks.} Summary of the
10 algorithmic optimization problems from ALE-Bench-Lite~\citep{imajuku2026ale}, drawn
from AtCoder Heuristic Contests, used to evaluate \modelname{}.}
\label{tab:algo_tasks}
\small
\renewcommand{\arraystretch}{1.15}
\begin{tabularx}{\linewidth}{@{}
  >{\raggedright\arraybackslash}p{0.29\linewidth}
  >{\raggedright\arraybackslash}p{0.07\linewidth}
  >{\raggedright\arraybackslash}X
@{}}
\toprule
\textbf{Problem / Category} & \textbf{Count} & \textbf{Description} \\
\midrule
ahc008 (Territory) & 1 &
Build partitions to maximize humans' reachable space while
reducing the number of pets sharing it. \\
\addlinespace[4pt]

ahc011 (Sliding Puzzle) & 1 &
Slide tiles to form the largest tree; minimize moves once
all tiles are connected. \\
\addlinespace[4pt]

ahc015 (Candy Clustering) & 1 &
Tilt a box to cluster candies by flavor, maximizing the
sum of squared cluster sizes. \\
\addlinespace[4pt]

ahc016 (Graph Classification) & 1 &
Design graphs robust to edge noise and vertex relabelling,
balancing classification errors against graph size. \\
\addlinespace[4pt]

ahc024 (Map Compression) & 1 &
Minimize occupied map cells while preserving district
connectivity and all adjacencies, including the exterior. \\
\addlinespace[4pt]

ahc025 (Weight Balancing) & 1 &
Partition items of unknown weight using limited balance
queries; minimize variation in group weights. \\
\addlinespace[4pt]

ahc026 (Box Stacking) & 1 &
Rearrange box stacks to remove boxes in increasing label
order with minimum relocation energy. \\
\addlinespace[4pt]

ahc027 (Cleaning Route) & 1 &
Find a closed route covering every grid cell to minimize
average dirt under repeated traversal. \\
\addlinespace[4pt]

ahc039 (Fishing Net) & 1 &
Enclose mackerels while avoiding sardines using a simple
rectilinear polygon under perimeter and vertex limits. \\
\addlinespace[4pt]

ahc046 (Skating with Blocks) & 1 &
Visit targets in order with minimum actions using steps,
slides, and block placement or removal. \\
\bottomrule
\end{tabularx}
\end{table}

\section{Model Pricing}
\label{app:model_pricing}

We present the token prices for the two model configurations
used in our experiments: GPT-5.6-Terra with GPT-5.6-Luna, and
GLM-5.3 with GLM-5.3-Flash in~\cref{tab:model_pricing}.

\begin{table}[hb]
\centering
\caption{\textbf{Model pricing.} Prices were retrieved from the official OpenAI website and
OpenRouter on September 4, 2026, for the GPT and GLM models,
respectively. All prices are in USD per million tokens.}
\label{tab:model_pricing}
\small
\renewcommand{\arraystretch}{1.15}
\begin{tabular}{@{}lrrrr@{}}
\toprule
\textbf{Model} & \textbf{Input} & \textbf{Output} &
\textbf{Cache Read} & \textbf{Cache Write} \\
\midrule
GPT-5.6-Terra & \$2.00 & \$12.00 & \$0.20 & \$2.50 \\
GPT-5.6-Luna  & \$0.20 & \$1.20  & \$0.02 & \$0.25 \\
\midrule
GLM-5.3       & \$1.15  & \$3.50 & \$0.10  & --- \\
GLM-5.3-Flash & \$0.075 & \$0.25 & \$0.015 & --- \\
\bottomrule
\end{tabular}
\end{table}

\section{Prompts}
\label{app:prompts}
Below are the prompts used for strategy exploration and code implementation and refinement.

\begin{promptbox}{Strategy Exploration Prompt}
    \begin{Verbatim}[breaklines=true]
You are the strategy designer for a program-evolution system.
Propose exactly {K} distinct strategies to improve the current
program's evaluated score.

Inputs:
- Task specification and validity constraints
- Current program and evaluated metrics
- Evaluator-aware guidance, when available
- Previous strategies, implementation results, and failures

Each strategy must describe:
1. Mechanism family: a short identifier for the approach.
2. Replacement scope: the function, phase, or block to replace.
3. Compute budget: explicit limits on additional computation.
4. Gain mechanism: the bottleneck addressed or new degree of
   freedom introduced.

Propose different mechanisms, not variations in random seeds,
tolerances, or iteration counts. For K >= 3, include local
refinement, subsystem replacement, and structural exploration.
Use previous results to avoid repeating exhausted approaches.

Make every plan self-contained and immediately implementable.
Preserve public interfaces and hard validity constraints.
Do not rely on unavailable tools, fabricated data, or an
unspecified future optimization step.

Explain why the proposed change can improve the actual scoring
objective. Compare quantities on the same scale. End each plan
with a concise gain check: either a justified score estimate
or a testable hypothesis tied to a specific new mechanism.
A fallback or an accept-only-if-better rule is not itself
evidence of improvement.

Return each strategy in this format:
<STRATEGY_TITLE>[mechanism-family] short title</STRATEGY_TITLE>
<STRATEGY>
Mechanism family: ...
Replacement scope: ...
Compute budget: ...
Gain mechanism: ...
Implementation plan and expected-gain check.
</STRATEGY>

When requested, also return one complete, executable portfolio
reference program in <REFERENCE_IMPLEMENTATION> tags.
It is an independent candidate, not assigned to any strategy.
Keep program code outside the strategy blocks.
    \end{Verbatim}
\end{promptbox}
\begin{promptbox}{Code Implementation and Refinement Prompt}
    \begin{Verbatim}[breaklines=true]
You are the implementation model in a program-evolution system.
Implement the supplied strategy to improve the current program.

Inputs:
- Task specification and validity constraints
- Current parent program and evaluated metrics
- Strategy to implement
- Previous implementation feedback, when available

Implement the strategy faithfully. Replace the specified
mechanism and remove superseded code when safe. Do not
substitute an unrelated idea, make only a cosmetic change,
or append another complete optimization pipeline.

Ensure that the new mechanism executes on the normal return
path. Preserve public interfaces, permitted edit boundaries,
and hard validity constraints. Keep added computation within
the strategy's budget. A valid fallback may be retained, but
returning only the fallback does not implement the strategy.

If the parent already implements the proposed mechanism,
refine that mechanism rather than adding a duplicate.
Check dimensions, data types, mutability, API calls, and
consistency between function definitions and call sites.

For a pilot implementation, build the smallest executable
version that exercises the strategy's distinctive mechanism.

For subsequent refinement attempts, use evaluated metrics and
errors to repair failures or revise unsuccessful choices.
Do not repeat failed patches or merely change random seeds.
For timeouts, reduce work or replace the expensive stage while
preserving the strategy's gain mechanism.

Return code in the required output format. For SEARCH/REPLACE
edits, copy each SEARCH block exactly from the current parent
and ensure that it matches uniquely. Previous attempts do not
replace the parent as the edit target.

When repairing a supplied failing program, return a complete,
coherently corrected program rather than a diff against the
failed attempt.
    \end{Verbatim}
\end{promptbox}

\end{document}